\documentclass[11pt]{article}
\usepackage{eacl2017}
\usepackage{times}
\usepackage{url}
\usepackage{latexsym}
\usepackage{amsmath}
\usepackage{multirow}
\usepackage{url}
\usepackage{anyfontsize}
\usepackage{LIreduced}
\usepackage{amssymb,amsfonts}
\usepackage{graphicx}
\usepackage{color}
\usepackage{booktabs}
\usepackage[table]{xcolor}
\usepackage[round,semicolon,compress]{natbib}
\newcommand{\cf}[1]{\mbox{$\mathit{#1}$}}   
\newcommand{\R}{\mathbb{R}}

\eaclfinalcopy 

\title{Collaborative Training of Tensors\\ for Compositional Distributional Semantics}

\author{Tamara Polajnar \\
  Computer Laboratory \\
  University of Cambridge \\
  {\tt tamara.polajnar@cl.cam.ac.uk} \\
 }

\date{}

\begin{document}
\maketitle
\begin{abstract}
Type-based compositional distributional semantic models present an
interesting line of research into functional representations of
linguistic meaning. One of the drawbacks of such models, however, is
the lack of training data required to train each word-type
combination. In this paper we address this by introducing training
methods that share parameters between similar words. We show that
these methods enable zero-shot learning for words that have no
training data at all, as well as enabling construction of high-quality
tensors from very few training examples per word.  
\end{abstract}

\section{Introduction}
Multiple compositional distributional
semantic models have been proposed in the past several years. Most models are based on a vector
representation for words and a separate function that performs
composition on the vectors
\citep{mitchellvector,socher2012semantic,zanzotto2012disttree}. Another
line of investigation represents atomic types (mainly nouns and sentences) as vectors
but predicate types, for example adjectives and verbs, as functions
that act upon the atomic types or other functions
\citep{coecke2010mathematical,baroni2014frege,grefenstette2011experimenting,kartsadrqpl2014,polajnar2014reducing}.

\begin{figure}
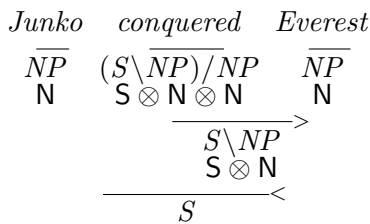

\begin{center}
\deriv{3}{
  \it Junko & \it conquered & \it Everest \\
  \uline{1} & \uline{1} & \uline{1}\\
  \cf{NP} & \cf{(S\backslash NP)/NP} & \cf{NP}\\
  \mathsf{N} & \mathsf{S} \otimes \mathsf{N} \otimes \mathsf{N} & \mathsf{N}\\
  & \fapply{2}\\
  & \mc{2}{\cf{S\backslash NP}}\\
  & \mc{2}{\mathsf{S} \otimes \mathsf{N}}\\
  \bapply{3}\\
  & \mc{1}{\cf{S}}
}
\end{center}
\caption{The composition of a subject-verb-object phrase with CCG. The
  categories are also described by the corresponding tensors
  constructed from vector spaces  $\mathsf{S} \in \R^{S}$ and
  $\mathsf{N} \in \R^{N}$, where
  $\otimes$ is tensor product.}
\label{svocomp}
\vspace{-0.6cm}
\end{figure}

In the full form of the Categorial Framework (CF)
\citep{coecke2010mathematical}, each lexical item and type pairing is
represented by a tensor whose order is determined by the number of
atomic-type arguments in its category. If we use Combinatory
Categorial Grammar (CCG) as the basis for further discussion in this
paper, the noun phrase type is \cf{NP}, which is represented by a
first order tensor, i.e.,  a vector. An adjective category is \cf{NP/NP},\footnote{In CCG the adjective is correctly represented by \cf{N/N}; however since our datasets do not include determiners, here we make no distinction between \cf{N} and \cf{NP}.}  and is
modelled by a matrix (a second order tensor), and a transitive verb is
a third order tensor whose category is \cf{(S\bs NP)/NP}. The verb
category is interpreted as looking for a noun phrase to the right
(object), a noun phrase to the left (subject), and when these are
found the function results in a complete sentence \cf{S}. Since the
subject and the object are represented by vectors and the verb is a
tensor, the natural composition operation is tensor contraction, which
is equivalent to matrix multiplication for second order tensors. The
result of the composition between the tensor, the subject, and
object vectors is a vector representing a composed sentence (Fig.~\ref{svocomp}). 

While the type-based approach has strong theoretical grounding \citep{baroni2014frege} and
integrates well with Categorial grammars
\citep{coecke2010mathematical,maillard2014tensor}, there are two major practical
challenges. The first, which has recently been addressed by
\citet{Fried2015},  is that there are many type-lexeme combinations and some of the
types lead to high-order tensors. This results in a large number of
parameters that need to be estimated and stored. The second  challenge is that some
of the type-lexeme combinations are very rare, and, therefore, may
never be observed in the training data, leading to difficulties in accurate
parameter estimation for those types. 

In this paper, we propose two methods that address the second, and yet unsolved, challenge and
we evaluate them on both adjectives and transitive verbs. This work
improves on some of the shortcomings in the basic
implementations of the Lexical Function (LF) model
\citep{baroni2010nouns} and the contemporaneous Categorial Framework (CF)
\citep{coecke2010mathematical}.

\section{Related Work}

The type-based models have a solid theoretical grounding
\citep{coecke2010mathematical, baroni2014frege} neatly combining
mathematical category theory and categorial grammars like CCG with
linear algebra to represent the meaning of argument taking words as
functions modelled in vector space. The tensors in the CF provide a way of representing words, like  verbs, as functions that can recognise the prototypical combinations of features that are expected from their arguments. So, for example,  a function representing the verb {\em eat} would  not only encode that it expects animate subjects that consume edible objects, but ultimately the types of edible objects that make sense with particular subjects. This is different from most other methods \citep{mitchellvector,socher2012semantic,zanzotto2012disttree} where all words are represented as vectors, and combined with operators of various complexity, which themselves do not encode any semantics. 

Due to the parameter explosion problems mentioned in Section~1, the CF implementations have mostly been tested on restricted constructions such as adjective-noun or subject-verb-object phrases, while the neural-network-based approaches have been optimised for and tested on full-sentence tasks  \citep{socher2012semantic, kiros:15}. By addressing the problem of training data sparsity, we hope to bring the CF a step closer to full implementation, so that it can be compared to other models on a variety of tasks. 

Various approximations of CF which curb the number of parameters have
previously been implemented
\citep{grefenstette2011experimenting,paperno2014practical,polajnar2014reducing},
but most of those changed the shape of the tensors in some way that
diminishes the spirit of the full model.  Of these, the Practical
Lexical Function (PLF) model of \citet{paperno2014practical} comes
closest to full sentence implementation of a type-based semantic
model.  It extends the type-based approach to full sentences by
representing argument taking types as a set of matrices, each of which
interacts with one argument. While this has the effect of reducing the
number of parameters, it also decouples the interactions between
arguments, which is one of the main strengths of the tensor-based
model.  In contrast, \citet{Fried2015} provide a mathematically
principled way of achieving parameter reduction while preserving the
shape of tensors, and hence the interactions between the arguments, by
employing tensor decomposition. As it neatly solves
the parameter explosion problem, we experimentally demonstrate how
this approach complements the methods introduced in this paper. 

The lack of training data has been partially addressed in
\citet{grefenstette2013multistep} by training third-order verb tensors
in two steps in order to take advantage of more plentiful verb-object
training data. On the other hand,  \citet{Polajnar2015} show that it
is possible to train  full verb tensors with a single-step
multi-linear regression method which is the basis of the approaches
described in this paper.  

\begin{figure}
\begin{center}
\begin{minipage}{5cm}
\includegraphics[width=\textwidth]{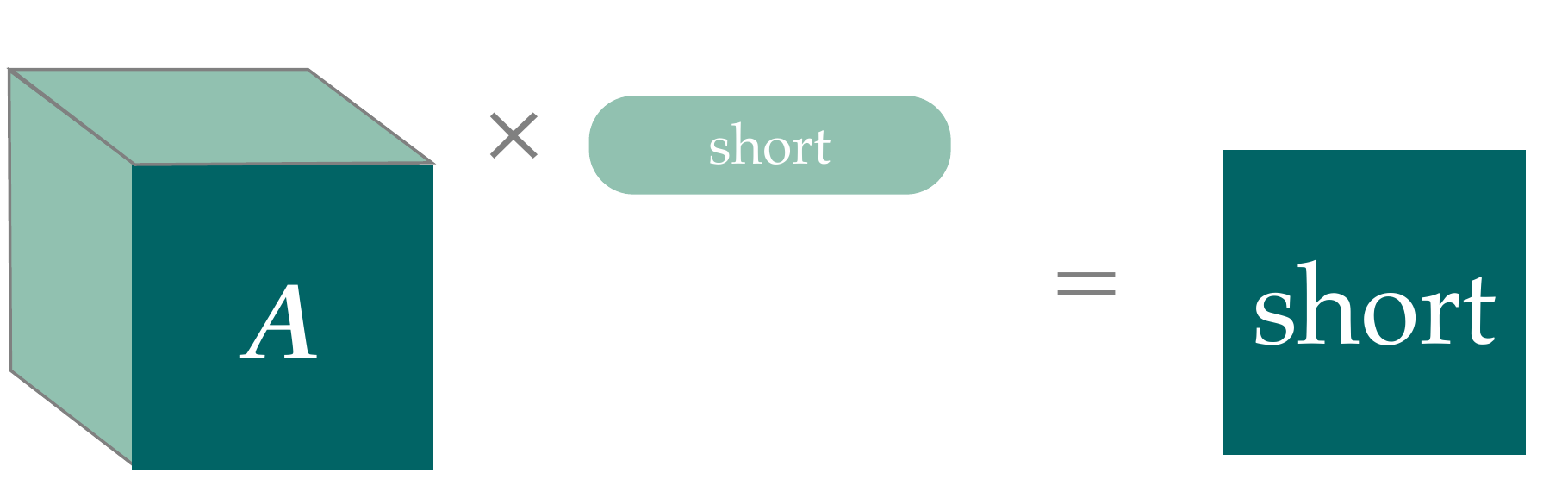}
\end{minipage}
\end{center}
\caption{Generalised Lexical Function (GLF).}
\label{fig:glf}
\vspace{-0.5cm}
\end{figure}

\citet{Bride2015} introduce the Generalised
Lexical Function (GLF), a method for overcoming the sparsity of
training data for adjectives. We reimplement this method for
comparison in this paper for the adjective case.  
GLF is trained in two steps. Firstly, for adjectives that have sufficient
training data matrices are obtained using the standard LF regression-based
training method. In the second step, the pre-trained matrices are used
along with vector representations of adjectives to train a third-order
tensor. When a vector for a new (test) adjective is
multiplied with this tensor, it should produce a matrix representing
that adjective (Fig.~\ref{fig:glf}). 

One drawback of this method is
that the LF training data and the adjective vectors have to come from
the same corpus. Therefore GLF requires a corpus with sufficient coverage to provide vector
representations for a full lexicon of adjectives,
where for our methods we can source new adjectives from alternative
corpora or ontologies. 

The second drawback is that it is only defined for adjectives.  If it 
was extended to transitive verbs, for example, this theoretical
extension would require training a fourth-order tensor which
generates third-order verb tensors. In contrast, our methods are easily
extensible to any word type, which we demonstrate by also applying them to
verbs, and only introduce
a scalar parameter each. 



\section{Background}

In order to build a semantic representation of  a word we need to see that
word in a variety of contexts many times. For most methods that build
vector representations it is enough to see a word used say $q$
times, where $q$ is some number which either empirically or
qualitatively leads to reasonable word vectors.

For type-based methods we require more complex training
data. For example, to train the adjective {\em red} we may need $p$ distinct
training examples of the word {\em red} being used in text as an
adjective, e.g. {\em red car, red velvet, red flower}. For each of
those examples we need a vector representation for the noun ({\em car}) and the {\bf holistic vector} for the noun phrase
({\em red car}). To get good quality representations for the holistic
vectors we need to see the exact phrase {\em red car} at least $q$
times. 

To train verbs the task is even more difficult as we need recurring three word phrases,
e.g. {\em Jones knits jumpers} which appear at least $q$ times with
nouns that also individually appear at least $q$ times. 

There are several ways that even relatively well represented
predicates can lead to poor training data. For example, we may see the word {\em crimson} in a corpus enough times to
build a word vector; however it may appear with a different
noun  each time, leading to poor quality training examples for the type-based
model. 

In this paper we describe two methods that can be used individually or
together to improve the representations of adjective and verb tensors
which have scarce training data. Both methods leverage knowledge that
two predicates are similar. This knowledge can be gathered from vector
similarities, as we do in this paper, because as we have seen we can
have a  good vector representation  from an unparsed corpus and with much less training data. Alternatively
similarity scores can be obtained from a thesaurus. 

\section{Methods}

In this section we first describe the methods used to train second and
third-order tensors as function representations of adjectives and
transitive verbs respectively. We also consider the low-rank
representations \citep{Fried2015,Fried2015thesis}, which were
demonstrated to produce competitive performance with fewer
parameters. We then extend these methods to include parameter sharing. 
All methods have been implemented using gradient descent, rather than analytic regression. 

\subsection{Basic Tensor Training}

\subsubsection{Adjective}

We model each adjective as a linear function that maps a noun to an adjective-noun phrase. The $N$-dimensional noun vectors ($\mathbf{n}$) are transformed into  $N$-dimensional noun phrase vectors via an $N\times N$ matrix $\mathbf{A}$, just as in the Lexical Function (LF) model of \citet{baroni2010nouns}. The loss function consists of minimising the error between the vector resulting from the adjective-noun multiplication and a holistic vector representing the adjective-noun phrase ($\mathbf{z}_{an}$):

\begin{equation}
    \label{eqn:adj_loss}
    L(\mathbf{A}) = ||\mathbf{A}\mathbf{n} - \mathbf{z}_{an}||_2
\end{equation}

\paragraph{Low-rank adjectives} 
We learn low-rank adjective matrices by fixing a maximal rank $R$ and maintaining each matrix in a rank-decomposed form, which is similar to the singular value decomposition (SVD) \citep{Fried2015thesis}. The low-rank representation for an adjective $\mathbf{A}$ is
\begin{equation}
    \mathbf{A} = \sum_{r=1}^R \mathbf{U}_r \otimes \mathbf{V}_r
\end{equation}
where $\mathbf{U} \in \R^{R \times N}, \mathbf{V} \in \R^{R \times N}$ are parameter matrices, $\mathbf{U}_r$ gives the $r$th row of matrix $\mathbf{U}$, and $\otimes$ is the tensor product.

The adjective matrix's action on vectors is then given by 
\begin{equation}
    \mathbf{An} = \mathbf{U}^\top (\mathbf{Vn})
\end{equation}
Adjectives then consist of $2\times NR$ parameters instead of $N\times N$.

\begin{figure*}[ht]
\begin{minipage}{3cm}
\includegraphics[width=\textwidth]{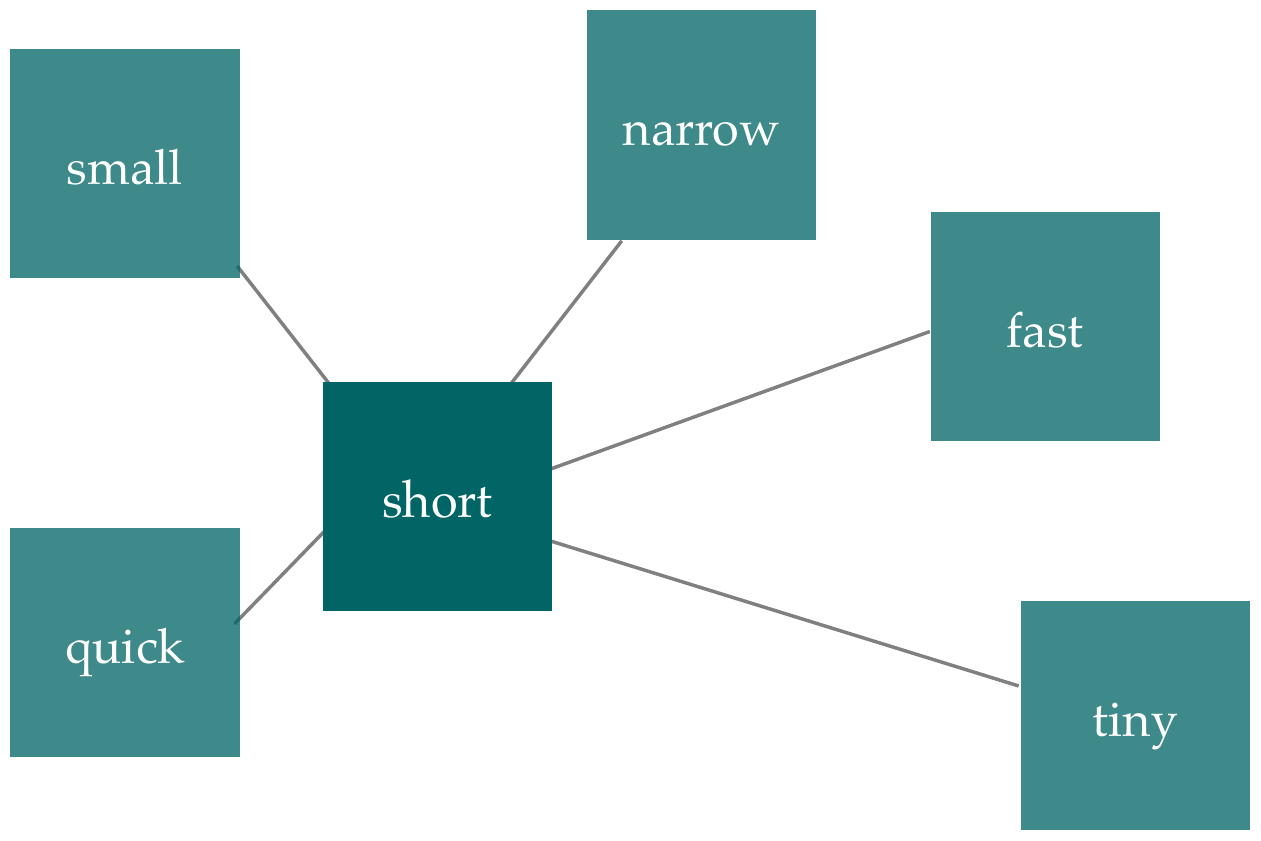}
\end{minipage}
\hspace{1cm}
\begin{minipage}{5cm}
\includegraphics[width=\textwidth]{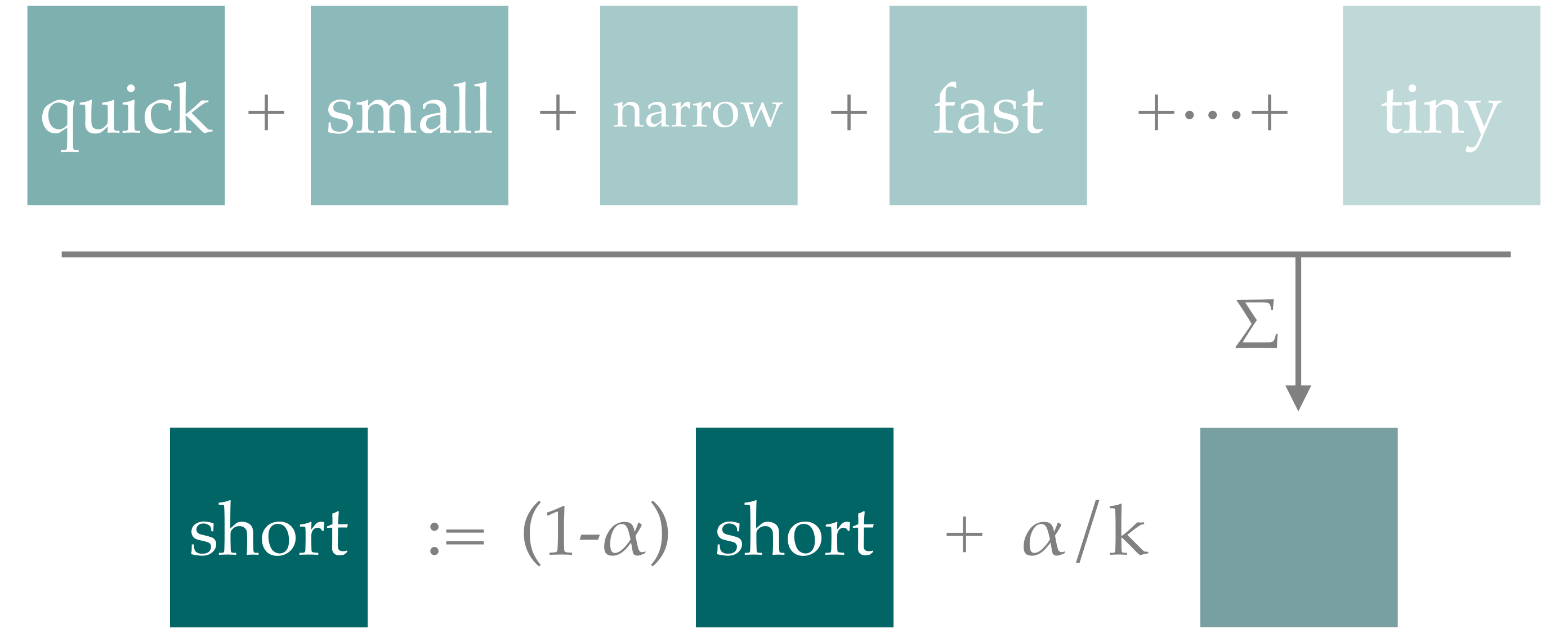}
\end{minipage}
\hspace{1cm}
\begin{minipage}{6cm}
\includegraphics[width=\textwidth]{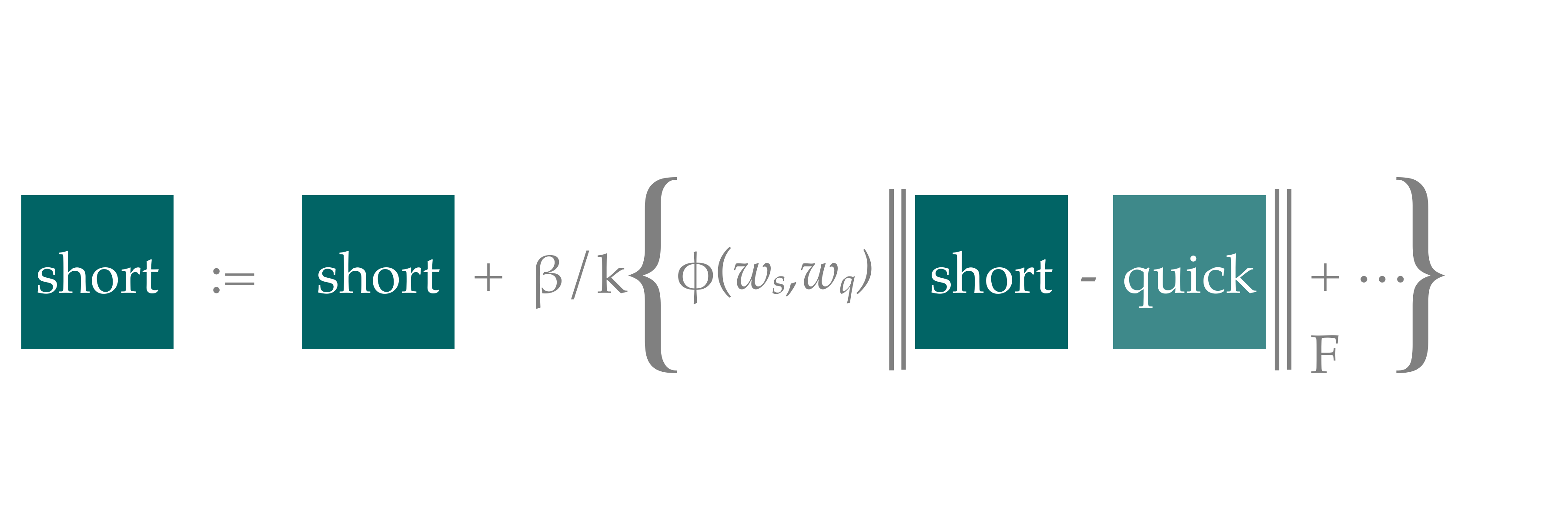}
\end{minipage}

\caption{From left, the similarity function $\phi$, Parameter Sharing
  (PS) where $\phi$ determines the intensity with which other tensors
  are added to the word that is being trained, and Fitting (FT) where
  $\phi$ regulates the distance between tensors directly.}
\label{fig:psft}
\vspace{-0.5cm}
\end{figure*}

\subsubsection{Verb}

We model each transitive verb as a bilinear function mapping subject and object noun vectors, each of dimensionality $N$, to a single sentence vector of dimensionality $S$.  Each transitive verb $V$ is associated with a third-order tensor $\mathcal{V} \in \R^{S\times N \times N}$, which defines this bilinear function. If vectors $\mathbf{n}_s \in \R^N$, $\mathbf{n}_o \in \R^N$ for subject and object nouns, respectively, then the loss function for each verb is:
\begin{equation}
    \label{eqn:verb_loss}
    L(\mathcal{V}) = ||\mathbf{n}_s\mathcal{V}\mathbf{n}_o - \mathbf{z}_{s}||_2
\end{equation}

That is the error between the sentence vector produced by applying
\emph{tensor contraction} (the higher-order analogue of matrix
multiplication) to the verb tensor and two noun vectors and the
distributional (holistic) representation for sentence $\mathbf{z}_s$.

\citet{Polajnar2015} examine several different distributional sentence spaces; from these we chose the intra-sentential contextual sentence space consisting of content words that occur within the same sentences as the SVO triple, disregarding the verb itself. 

\paragraph{Low-rank verbs} Following \citet{Fried2015}, we use {\em canonical polyadic (CP) decomposition} representation of verb tensors. 
CP decomposition factors a tensor into a sum of $R$ tensor products of vectors, reducing the number of parameters we have to learn.

The low-rank representation for a verb $\mathcal{V}$ is: 
\vspace{-0.3cm}
\begin{equation}
    \label{eqn:decomp}
    \mathcal{V} = \sum_{r=1}^R \mathbf{P}_{r} \otimes \mathbf{Q}_{r} \otimes \mathbf{R}_{r}
\end{equation}
where $\mathbf{P} \in \R^{R \times S}, \mathbf{Q} \in \R^{R \times N}, \mathbf{R} \in \R^{R \times N}$ are parameter matrices. 

Representing tensors in this form allows us to avoid explicitly generating the full tensor by formulating the verb tensor's action on noun vectors as matrix multiplication:
\vspace{-0.3cm}
\begin{equation}
    \label{eqn:low_rank_contraction}
    V(\mathbf{s}, \mathbf{o}) = \mathbf{P}^\top (\mathbf{Qs} \odot \mathbf{Ro})
\end{equation}
where $\odot$ is the elementwise vector product. As a result we are
able to reduce the number of parameters needed to model each verb from
$S \times N \times N$ to $R \times (2N + S)$.





\subsection{Collaborative Tensor Training}

While low-rank methods reduce the amount of memory required to store
the parameters for a full lexicon and the amount of time required to train the tensors, they do not address the problem of data sparsity. To train tensors we need high-quality examples, which have to be extracted from parsed data. However, there are predicates for which there are few reliable training examples and others for which there are no training examples at all. In those cases we would still like to have a non-zero approximation for the particular function. 

We propose two approaches to address these instances of sparsity
(Fig.~\ref{fig:psft}). Both approaches are based on the existence of
an external method that gives similarity between the words for which
we are trying to build tensors. We define this as  a function
$\phi(w_1, w_2)$ which gives us the similarity between words $w$
corresponding to tensors $\mathbf{T}_1$ and $\mathbf{T}_2$,  where
$\mathbf{T}$ refers to either adjective matrices or verb third-order
tensors as the approaches are analogous across the types. This can be
any method that provides a distance between words,  whether it is
manual or derived from an ontology or any distributional or
distributed representation of these words. If the method relies on
vectors, then these do not have to match the training data at all, as
we only rely on a matrix of similarity values between all pairs of
adjectives (or separately verbs) that we are training.  We only use the $K$ top most similar tensors according to $\phi$, where $K$ is currently a manually chosen parameter. 




The first approach (PS) shares parameters between tensors that we have
declared to be similar by directly creating a weighted average of the
target tensor with the sum of the tensors of the $K$ closest words
(weighted by the similarity values from $\phi$).  As we iterate
through the gradient descent algorithm the parameters from the tensors
that have training data gradually disseminate and blend to produce
unique representations for words without training data.  

The second approach (FT) is similar to retrofitting \citep{Faruqui15}
and uses a regularisation component to push a
tensor closest to its nearest neighbours (according to $\phi$) by
encouraging smaller distances between them. In the experimental
sections we apply these two methods individually and together. 
Both methods are regulated by parameters and thus can be used
separately or jointly. 



\subsubsection{Parameter Sharing (PS)}
In this first method we share parameters between most similar tensors
using the function $\phi$. We adjust the appropriate loss function
(Eq.~\ref{eqn:adj_loss} or Eq.~\ref{eqn:verb_loss}) to incorporate
parameter sharing (PS) during gradient descent, e.g. for adjectives:

\vspace{-0.2cm}

{\small
\begin{equation} 
\label{eqn:ps_loss}
    L_{ps}(\mathbf{A}) = \left|\left|\left[(1-\alpha)\mathbf{A} +
      \frac{\alpha}{K}\sum_{i=1}^K \phi(w, w_i) \mathbf{A}_i\right]\mathbf{n} - \mathbf{z}_{s}\right|\right|_2
\end{equation}
}
\vspace{-0.2cm}

The parameter $\alpha$ balances the amount of tensor we are replacing
by the average of the nearby tensors. In case of the low-rank representations of tensors,  we use the deconstructed versions of the tensors and share the parameters between corresponding decomposed matrix representations by aligning the $\mathbf{U}$, $\mathbf{V}$, and for verbs $\mathbf{W}$, for the word pairs without reconstructing the tensors. 

\subsubsection{Fitting (FT)}

The second method is used in place of $l_2$-regularisation to push the
parameters of the current tensor closer to the parameters of the
tensors of the similar words. The regularisation component is
\vspace{-0.3cm}
\begin{equation}
\label{eqn:ft_reg}
    R_{ft}(\mathbf{T}) = \frac{\beta}{K}\sum_{i=1}^K \phi(w, w_i) \||\mathbf{T} - \mathbf{T}_i||_F
\end{equation}
\vspace{-0.2cm}

\noindent and is integrated into the training function via the parameter $\beta$:
\begin{equation}
\label{eqn:ft_loss}
    L_{ft}(\mathbf{T}) = L(\mathbf{T}) + R_{ft}(\mathbf{T}) 
\end{equation}
Like with PS, in the low-rank representations we regularise each of the component matrices separately.







\section{Experimental Settings}
\subsection{Test Datasets}
We use several datasets that test composition or directly compare the quality of the produced tensors:

\noindent{\bf ML10:} adjective-noun (AN) pairs rated for similarity 
 \citep{Mitchell10}.\\
{\bf MEN:} word-word pairs rated for relatedness from which we extract the adjective-adjective pairs only \citep{Bruni14}.\\
{\bf SIMLEX:} word-word pairs rated for similarity from which we extracted the adjective-adjective and verb-verb pairs \citep{Hill14}. \\
{\bf GS11:} a verb disambiguation dataset consisting of subject-verb-object (SVO) triples arranged in pairs, where in each pair the subject and the object remain the same but the verb changes \citep{grefenstette2011experimenting}.\\
{\bf KS14:} a dataset subject-verb-object sentence pairs rated for similarity \cite{kartsadrqpl2014}, which is an extension of the verb-object component of the ML10 dataset. \\
{\bf ANVAN:} a verb disambiguation dataset containing pairs of adjective-noun-verb-adjective-noun sentences where only the verb varies \citep{kartsaklis2013sep}.  

\subsection{Training Data}
In order to train the tensors for the adjectives and verbs occurring in the above test data we need to find examples of their usage in text. We use the October 2013 dump of Wikipedia articles, which was tokenised using the Stanford NLP tools,\footnote{\url{http://nlp.stanford.edu/software/index.shtml}}
lemmatised with the Morpha lemmatiser \citep{carroll:01}, and parsed with the C\&C parser \citep{clark:cl07}. 

We use the parser output to find adjective-noun and
subject-verb-object combinations (tuples) that involve our target words. From
these we choose up to 500 tuples that occur at least twice ($p$), and that contain nouns that occur at
least 100 times ($q$). Some words are quite rare and do not have any training data, e.g. the adjective {\em ashamed}, or very little training data, e.g. adjectives {\em glad} and {\em gritted}, each of which has a single training example.

The vectors for nouns and the holistic vectors for the AN and SVO
phrases are generated using the Paragraph Vector
\citep{le2014distributed} model.
\footnote{\url{https://groups.google.com/d/msg/word2vec-toolkit/Q49FIrNOQRo/J6KG8mUj45sJ}}

\subsection{Model Training}

The LF model forms the basis of all the collaborative training models. The adjectives and verbs are trained up to 200 iterations using batched gradient descent with ADADELTA \citep{zeiler2012adadelta}, at which point most of the tensors have finished training. The stopping criterion for adjectives is stagnation or an increase in training error. For verbs we also use a 10\% validation dataset if there are at least 20 training points. Full tensor training also uses $l_2$ regularisation with parameter 0.1, while sparse tensors are trained without regularisation as this was observed to be more optimal in \citet{Fried2015thesis}.  


\section{Adjective Experiments}
We use ML10 as a development dataset and test a range of mixing
parameters for PS and FT, and how they work together.  We test the
effects of  training data 
sparsity through two experiments. To train adjectives we require
training data consisting of vector tuples ({\em noun, noun\_phrase}), e.g. ({\em
  red, red\_car}). 

In {\bf EX1} we examine the case where we have minimal training data. We vary the percentage of training tuples per tensor between 1\% and 100\%.
100\%  represents the maximum of 500 tuples, although some tensors
have fewer tuples. 

In {\bf EX2} we examine the case  where we have no training data for
particular adjectives. 
We simulate the complete lack of training data by keeping the
training tuples for a percentage of adjectives and ignoring the rest.
We vary the percentage of adjectives with training data from 1\% to
100\% (where  $100\% = 297$ adjectives).

We can combine both methods (PS, FT) with the lexical function 

\vspace{-0.4cm}
\begin{eqnarray*} 
\label{eqn:ps_loss}
    L(\mathbf{T}) =  L(\mathbf{T})_{ps} +    R_{ft}(\mathbf{T})
\end{eqnarray*}
\vspace{-0.2cm}

and tune the PS parameter $\alpha$ and FT parameter $\beta$ across a range of values ($\alpha \in
\{0, 0.1, 0.5, 0.9, 1\}$, $\beta \in \{0, 0.01, 0.05, 0.1\}$) to find
the best performing parameters on ML10. 

On full tensors  we found that it is sufficient to use one method or
the other, with PS being in general more robust to sparsity of both
kinds. In EX1 smaller $\alpha$
values work well when there are few training tuples per adjective but
performance of the 
larger values increases along with the available training data. In EX2 $\alpha =0.9$ was consistently dominant. Variations in $\beta$ provided small
improvements of 2-3\% over the baseline PS performance. Based on these
conclusions we chose two settings for the experiments:  PS+FT$_{fix_1}$ with
$\alpha=0.9$, $\beta=0.01$ and PS+FT$_{var}$ where $\alpha$ increases
from 0 to 0.9 with the number of training tuples, so in  EX2, PS+FT$_{var}$ is equivalent to PS+FT$_{fix_1}$.


On decomposed tensors, the smaller number of parameters leads to more
volatile performance.  We ranked the performance of all $(\alpha,
\beta)$ pairs and chose two fixed settings corresponding to the two highest ranking parameter pairs $(0, 0.1)$,  PS+FT$_{fix_2}$, and $(0.1, 0.1)$, PS+FT$_{fix_3}$. The former setting is just FT as parameter sharing is null.

The greyed out boxes in Table~\ref{table:exp1res} show the values that
were obtained in tuning with these settings.


\begin{table*}[!t]
{\scriptsize
\begin{center}
\begin{tabular}{|l|c c c c c|c c c c c|c c c c c|}
\multicolumn{16}{c}{\bf Adjective Experiment 1 (EX1)}\\\hline
{\bf Full Tensor} & \multicolumn{5}{c|}{\bf ML10} & \multicolumn{5}{c|}{\bf MEN} & \multicolumn{5}{c|}{\bf SIMLEX}\\ \hline
Method & 1\% & 5\% & 30\% & 70\% &100\% & 1\% & 5\% & 30\% & 70\% &100\% & 1\% & 5\% & 30\% & 70\% &100\% \\ \hline

LF & - & - & \textcolor{gray!45}{0.05} & \textcolor{gray!45}{0.04} &0.47 & - & - & \textcolor{gray!45}{ 0.08 }& \textcolor{gray!45}{0.05}  & 0.39 &-&-&\textcolor{gray!45}{-0.12}&\textcolor{gray!45}{-0.04}& \textcolor{gray!45}{0.10}\\
GLF & 0.19 & {\bf 0.47} &{\bf 0.49} & 0.46 & 0.47 & \textcolor{gray!45}{0} & 0.33 & {\bf 0.58} & {\bf 0.57} & 0.50  & \textcolor{gray!45}{0.13}&0.22& 0.34& 0.33&0.35\\
PS+FT$_{fix_1}$ &\cellcolor{gray!25} 0.24 &\cellcolor{gray!25} 0.40  & \cellcolor{gray!25} 0.43 & \cellcolor{gray!25} 0.46 &  {\bf 0.49} &  \textcolor{gray!45}{-0.11} &{\bf 0.38} & 0.29 & 0.33&{\bf  0.61} & \textcolor{gray!45}{0.01}&\bf0.47&\bf0.43&\bf0.45&\bf0.55\\ 
PS+FT$_{var}$ & \cellcolor{gray!25} {\bf 0.31}&\cellcolor{gray!25} 0.42 &\cellcolor{gray!25}0.44 &\cellcolor{gray!25} 0.46  & {\bf 0.49} & \textcolor{gray!45}{-0.08} & 0.36 & 0.31 & 0.33 & {\bf 0.61} &\bf0.39&0.45&0.40&{\bf 0.45}&{\bf 0.55}\\\hline

{\bf Low Rank} & \multicolumn{5}{c|}{\bf ML10} & \multicolumn{5}{c|}{\bf MEN} & \multicolumn{5}{c|}{\bf SIMLEX}\\ \hline

LF &-&-&\textcolor{gray!45}{0.05}&
\textcolor{gray!45}{0.04}&0.45&-&-&\textcolor{gray!45}{0.08}&\textcolor{gray!45}{0.04}&0.39
& - & -
&\textcolor{gray!45}{-0.12}&\textcolor{gray!45}{-0.05}&\textcolor{gray!45}{0.07}\\
PS+FT$_{fix_2}$&\cellcolor{gray!25} 0.25 & \cellcolor{gray!25} 0.12 & \cellcolor{gray!25} 0.37 & \cellcolor{gray!25} 0.42 &0.45 & 0.33 &\textcolor{gray!45}{-0.07}&\textcolor{gray!45}{0.14}& 0.52 & 0.54  &\bf 0.61&\bf 0.51&0.32&0.32&0.38\\
PS+FT$_{fix_3}$ &\cellcolor{gray!25} {\bf 0.39}&\cellcolor{gray!25} \bf 0.39 &\cellcolor{gray!25} 0.32 &\cellcolor{gray!25} 0.39 &0.47& 0.24& {\bf 0.40}&0.38&0.45& {\bf 0.62}  & 0.54& 0.50&\bf 0.43&\bf 0.45&\bf0.49\\\hline

\multicolumn{16}{c}{}\\
\multicolumn{16}{c}{\bf Adjective Experiment 2 (EX2)}\\\hline

 \bf Full Tensor & \multicolumn{5}{c|}{\bf ML10} & \multicolumn{5}{c|}{\bf MEN} & \multicolumn{5}{c|}{\bf SIMLEX} \\ \hline
Method & 1\% & 5\% & 30\% & 70\% &100\% & 1\% & 5\% & 30\% & 70\% &100\% & 1\% & 5\% & 30\% & 70\% &100\% \\ \hline
LF & 0.14 & 0.37 & 0.45 & 0.46 & 0.46 & 0.20 & 0.42 & 0.45 &0.39 & 0.39 & 0.24 & 0.20 & \textcolor{gray!45}{0.11} & \textcolor{gray!45}{0.09} & \textcolor{gray!45}{0.09}\\
GLF & 0.38 & 0.47 & 0.49 & 0.48 & 0.47 &  0.46  & 0.48 & 0.50 & 0.49 & 0.50 & 0.45 & 0.41 & 0.38 & 0.36 & 0.35\\
PS+FT$_{fix_1}$ &\cellcolor{gray!25} {\bf 0.52} &\cellcolor{gray!25} {\bf 0.52}  & \cellcolor{gray!25} {\bf 0.50} & \cellcolor{gray!25} {\bf 0.49} & {\bf 0.49} & {\bf 0.51} &{\bf 0.56} & {\bf 0.63} &{\bf 0.63} & {\bf 0.61} & \bf 0.68 & \bf 0.65 & \bf 0.58 & \bf 0.55 & \bf 0.55\\ \hline

{\bf Low Rank} & \multicolumn{5}{c|}{\bf ML10} & \multicolumn{5}{c|}{\bf MEN}  & \multicolumn{5}{c|}{\bf SIMLEX}\\ \hline
LF  &0.17&0.29&0.36&0.43&0.44&\textcolor{gray!45}{0}&0.28&0.41&0.40&0.38 & 0.10 & 0.11 & 0.03 & 0.07 & 0.07\\ 
PS+FT$_{fix_2}$&\cellcolor{gray!25} \bf 0.36  & \cellcolor{gray!25} 0.43 & \cellcolor{gray!25} \bf 0.46 & \cellcolor{gray!25} \bf 0.47  & 0.45 & \bf 0.42 & \bf 0.57 & \bf 0.65 & 0.58& 0.54 & 0.36 & 0.41 & 0.42 & 0.40 & 0.38\\
PS+FT$_{fix_3}$ &\cellcolor{gray!25} {\bf 0.36} &\cellcolor{gray!25} 0.43&\cellcolor{gray!25} 0.42 &\cellcolor{gray!25} 0.45  & \bf 0.47 &0.34& 0.46& 0.63& \bf 0.60 & \bf 0.62 & \bf 0.48 & \bf 0.47 &\bf  0.50 &\bf 0.51 & \bf 0.49\\\hline
\end{tabular}
\end{center}
}
\caption{Top: Spearman values as the percentage of adjectives with training
  data is increased (EX1). Bottom: Spearman values as the percentage of training data per adjective is increased (EX2). Grey cells indicate values were used as test data during parameter tuning, while grey numbers indicate non-significant correlations. Highest values in each section and column are bolded.}
\label{table:exp1res}
\end{table*}

\begin{table*}
{\scriptsize
\begin{center}
\setlength\tabcolsep{3px}
\bgroup
\def\arraystretch{1.2}
\begin{tabular}{|l|c c c c c|c c c c c|c c c c c|c c c c c|}
\multicolumn{21}{c}{\bf Verb Experiment 1 (EX1)}\\\hline
 \bf Full Tensor & \multicolumn{5}{c|}{\bf KS14} &
 \multicolumn{5}{c|}{\bf GS11}  &\multicolumn{5}{c|}{\bf ANVAN} & \multicolumn{5}{c|}{\bf SIMLEX} \\ \hline
Method & 1\% & 5\% & 30\% & 70\% &100\% & 1\% & 5\% & 30\% & 70\% &100\% & 1\% & 5\% & 30\% & 70\% &100\% & 1\% & 5\% & 30\% & 70\% &100\%  \\ \hline
Tensor & -&-&0.11&0.33&\bf 0.54 &-&-&-0.06&\bf 0.29&\bf 0.38
&-&-&0.07&\bf 0.10&0.14
&-&-&\textcolor{gray!45}{0.03}&\textcolor{gray!45}{0.02}&\textcolor{gray!45}{0.02}
\\  
PS+FT$_{fix_1}$ & 0.35&0.29&\bf 0.34&\bf 0.39& 0.50 & \bf 0.03&\bf 0.08&\bf 0.35&0.25&0.31 &0.11&0.04&0.08&0.07&\bf 0.18 &\bf 0.16&0.33&0.27&\textcolor{gray!45} {0.12}&\textcolor{gray!45} {0.12}\\ 
PS+FT$_{var}$ & \bf 0.44& \bf 0.37 &0.33&\bf 0.39& 0.50 &\textcolor{gray!45} {0.02}&0.05&\bf0.35&0.25&0.31 &\bf 0.14& \bf 0.06&\bf 0.10&0.07&\bf 0.18 &\bf 0.16&\bf 0.36&\bf 0.29&\textcolor{gray!45} {0.12}&\textcolor{gray!45} {0.12} \\ \hline

{\bf Low Rank} & \multicolumn{5}{c|}{\bf KS14} & \multicolumn{5}{c|}{\bf GS11}  &\multicolumn{5}{c|}{\bf ANVAN} & \multicolumn{5}{c|}{\bf SIMLEX} \\ \hline
Tensor & -&-&-0.04&0.25&0.52 &-&-&-0.08&0.23&0.35 &-&-&-&0.07&\bf 0.15 &-&-&\textcolor{gray!45}{0.01}&\textcolor{gray!45}{0.07}&\textcolor{gray!45}{0.08}\\ 
PS+FT$_{fix_2}$ & \bf 0.19& \bf 0.19&0.05&\bf 0.45&\bf 0.54 &\bf 0.25& \bf 0.26&0.17&\bf 0.24&\bf 0.44&\textcolor{gray!45} {-0.01}&\textcolor{gray!45} {0.01}&-0.07&\bf 0.13&0.12 &\textcolor{gray!45} {0.06}&0.19&\bf 0.18& \bf 0.17&\textcolor{gray!45} {0.09}\\ 
PS+FT$_{fix_3}$ & \textcolor{gray!45} {-0.03}&-0.11& \bf 0.17 & 0.26&0.38 &\textcolor{gray!45} {0.02}&\textcolor{gray!45} {0.02}&\bf0.24&0.16&0.23 &-0.05&\bf 0.02&\bf 0.09&0.11&0.10 &\bf 0.40&\bf 0.33&0.13&\textcolor{gray!45} {0.08}&\textcolor{gray!45} {0.01}\\\hline

\multicolumn{21}{c}{}\\
\multicolumn{21}{c}{\bf Verb Experiment 2 (EX2)}\\\hline
 \bf Full Tensor & \multicolumn{5}{c|}{\bf KS14} & \multicolumn{5}{c|}{\bf GS11}  &\multicolumn{5}{c|}{\bf ANVAN} & \multicolumn{5}{c|}{\bf SIMLEX} \\ \hline
Method & 1\% & 5\% & 30\% & 70\% &100\% & 1\% & 5\% & 30\% & 70\% &100\% & 1\% & 5\% & 30\% & 70\% &100\% & 1\% & 5\% & 30\% & 70\% &100\%  \\ \hline

Tensor & 0.17 &0.40&{\bf 0.53} & \bf 0.56 & \bf 0.54 & \bf 0.28& \bf 0.27 & \bf 0.35 & \bf 0.38 & \bf 0.39 &\textcolor{gray!45}{-0.02}&\bf 0.20&0.15& 0.14 & 0.16& \textcolor{gray!45}{-0.05} & \textcolor{gray!45} {0.03} & \textcolor{gray!45} {0.03} &\textcolor{gray!45} {0.03}&\textcolor{gray!45} {0.02}\\ 
PS+FT$_{fix_1}$ & {\bf 0.36} & {\bf 0.44} &0.49 & 0.50 & 0.49& 0.12 &0.18 & 0.25&0.26&0.29&\textcolor{gray!45}{0.02}&0.17&\bf 0.16&\bf 0.16 &\bf 0.18 & \bf 0.47 & \bf 0.18 & \textcolor{gray!45} {0.12} & \textcolor{gray!45} {0.13} &\textcolor{gray!45} {0.12}  \\ \hline  

{\bf Low Rank} & \multicolumn{5}{c|}{\bf KS14} & \multicolumn{5}{c|}{\bf GS11}  &\multicolumn{5}{c|}{\bf ANVAN} & \multicolumn{5}{c|}{\bf SIMLEX} \\ \hline

Tensor & 0.07 &0.25&\bf 0.46& \bf 0.47& 0.48& 0.20 & 0.21& 0.29 &0.35& 0.37&\bf 0.03&0.07&0.10& \bf 0.12 &\bf 0.16& \textcolor{gray!45}{-0.02} &\textcolor{gray!45}{-0.01}  & \textcolor{gray!45} {0.02} &\textcolor{gray!45} {-0.04}&\textcolor{gray!45} {-0.03}\\

PS+FT$_{fix_3}$& \textcolor{gray!45}{0}  &  0.14 & 0.14 & 0.23  & 0.30 & \textcolor{gray!45}{0.10} & 0.12 & \textcolor{gray!45}{-0.01} & 0.09 & 0.19 &\textcolor{gray!45}{-0.02} & 0.06 & -0.07 & 0.09 & 0.06 & \bf 0.25&\bf 0.24&\bf 0.28& \textcolor{gray!45}{-0.04} & \textcolor{gray!45}{-0.03} \\\hline

\end{tabular}
\egroup
\end{center}
}
\caption{Top: Spearman values as the percentage of verbs with training
  data is increased (EX1). Bottom: Spearman values as the percentage of training data per verb is increased (EX2). Grey numbers indicate non-significant correlations. }
\label{table:exp1resverb}
\end{table*}

 \subsection{Testing}



We test for adjective-noun composition (ML10), word-word relatedness (MEN) and similarity (SIMLEX). We compare the composed AN vectors and the unfurled adjective matrices using cosine similarity. 

Table~\ref{table:exp1res} reports the testing
results for EX1 and EX2 with the above settings and compare them with
GLF and LF. In the grey cells we have the values which were considered
when choosing the parameter settings. The rest of the values in the
tables represent testing results without tuning on these datasets. The
light grey numbers are not significantly correlated with the data with
$p < 0.05$. Some negative correlations occur in the testing results
(see  Sec.~\ref{qual}).

In Table~\ref{table:exp1res}, the columns that are marked with 70\% or less on all datasets show that LF requires at least some training data for each adjective.  GLF performs well when there is moderate sparsity in the percentage of available adjectives (5\%-70\%). PS+FT gets highest results on MEN and SIMLEX more often than GLF, and most significantly both GLF and PS+FT regularly outperform LF even when all the training data is available.

ML10 additive baseline is 0.52, and the vector similarity baseline is 0.48 for MEN and 0.37 for SIMLEX on 100\% of the data.  The additive baseline on ML10 with the word2vec vectors is difficult to beat, although PS+FT does match it occasionally. More interestingly the collaborative representations do better on word-word relatedness (MEN) and similarity (SIMLEX) than both word vectors and LF. LF does not reach statistically significant correlations on SIMLEX even with 100\% of training data, while the collaborative methods do overwhelmingly better.

\section{Verb Experiments}
We repeat ablation testing  with the 285 verbs, in order to see if our
methods are effective in situations where there is lack of training
data. These verbs occur in the
datasets that cover: SVO relatedness (KS14), verb disambiguation
within a noun-verb-noun (GS11) and an
adjective-noun-verb-adjective-noun SVO contexts (ANVAN), and verb-verb
similarity (SIMLEX). The top part of Table~\ref{table:exp1resverb}
shows results as the number of verbs with training data is increased
(EX1) and the bottom part shows results as the number of training
examples per verb is increased (EX2). The additive baseline results on
100\% of the data are 0.59 for KS14, 0.13 for GS11, 0.03 for ANVAN,
and the vector similarity baseline is 0.14 for SIMLEX.  We see large
improvements over the low GS11, ANVAN, and SIMLEX baselines; however,
the additive baseline on KS14 which is an extension of the ML10
verb-object dataset is difficult to beat. This is constant with
previous findings, as are the scores we achieve here
\citep{Fried2015}.  We are not trying to produce the best performance,
but instead our experiments show that these methods work
when there is lack of data.

Since we did not retune our methods specifically for tensors, we see greater variability in performance, with the original tensor model often outperforming the collaborative training methods when there is enough data available. The introduced methods are still useful when there is little or no training data. Surprisingly, we see higher performance from  low-rank approximations of tensors, indicating that they may be more stable than the approximations of matrices. 

\begin{table}
{\scriptsize
\begin{center}
\begin{tabular}{|c c||c c|} 
\multicolumn{2}{|c||}{\bf yellow} & \multicolumn{2}{c|}{\bf play} \\ \hline
\multicolumn{1}{|c}{\bf PS+FT} & \multicolumn{1}{c||}{\bf LF} & \multicolumn{1}{c} {\bf PS+FT} & \multicolumn{1}{c|}{\bf Tensor} \\ \hline
orange & red & participate & make \\
red & blue & make & start \\
blue & white & start & do  \\
coloured & green & do & join \\
brown & brown & win & get  \\ \hline
\multicolumn{2}{|c||}{\bf outdoor} & \multicolumn{2}{c|}{\bf entangle} \\ \hline
\multicolumn{1}{|c}{\bf PS+FT} & \multicolumn{1}{c||}{\bf LF} & \multicolumn{1}{c} {\bf PS+FT} & \multicolumn{1}{c|}{\bf Tensor} \\ \hline
 domestic & many & win-over & transmit \\
 local & large & attach & knock \\
outer & various & disorganize & deposit \\ 
new & new & multiply  & eat \\
foreign & small & tap & wash \\ \hline
\end{tabular}
\end{center}
\caption{Nearest neighbours of two adjectives and two verbs with and without collaborative training.}
\label{nnex}
}
\vspace{-.5cm}
\end{table}

\section{Qualitative Analysis}
\label{qual}

Since the goal of training tensors for the Categorial Framework is good performance in composition tasks, we tuned our parameters on an adjective-noun composition dataset (ML10). In Table~\ref{table:exp1res} we can see that when we have only 2 adjectives with training data (1\% column) we have acceptable performance on this dataset; but,  the performance on the datasets where we compare the adjective matrices directly (MEN and SIMLEX) is statistically non-correlated and sometimes negative. Nearest neighbours analysis of adjectives produced by the parameter settings $fix_1$ and $var$ shows that similarities between all adjectives approach one and lead to nonsensical rankings. This is due to the fact that the numbers in the matrices themselves are very low and approach machine precision. Using cosine similarity leads to elementwise multiplication between low numbers and hence near-zero values in the numerator. On the other hand, composition is performed using matrix multiplication between the adjective matrices and the noun vectors, which have much larger numbers, resulting in the more sensible performance in the composed datasets. An alternative way of evaluating the quality of the tensors  would be to treat them as functions, and instead of cosine employ the function comparison within the type-driven framework introduced by \citet{Maillard15}.

Another interesting phenomenon we observed is that PS+FT works better than LF when all adjective training data is available. So if we compare the nearest neighbours for adjectives produced by $fix_1$ setting to the ones produced with LF, we can notice qualitative improvement (Table~\ref{nnex}). Keeping in mind that our pool of nearest neighbours is limited to the 297 adjectives in our training data,  we can see subtle differences in the adjective {\em yellow},  which is well represented with a large amount of training data and nearest neighbours. Although all of the closest terms are chromatic,   {\em orange} and {\em red} are the closest when collaborative training is involved. For the underrepresented adjective {\em outdoor} we can see that PS+FT finds more semantically related and less general adjectives although true neighbours are not available. 

For verbs we found that PS+FT often does worse than the straight-forward tensor method (Table~\ref{table:exp1resverb}). In Table~\ref{nnex}, we can see an example of easy to train verb {\em to play} where PS+FT does indeed rate a similar term {\em participate} highly. In contrast, the verb {\em to entangle} is rarer, and hence would have less training data and poorer vector representation.  
The closest term is {\em win-over}, a verb which is artificially hyphenated in the ANVAN dataset and hence has no naturally occurring training data.  
Together adjective and verb results indicate that a larger training pool from which we can choose related tensors may produce a better representations.

\section{Discussion}

In this paper we introduced two methods that address the lack of training data in the type-driven framework for compositional distributional semantics. In our experiments we use distributed vectors which have been found to achieve state-of-the-art results on some of the datasets we used here \citep{Fried2015}; however, the goal here was to compare these methods to the standard regression approach where each tensor is trained separately. We find that for both full tensors and low-rank approximations collaborative training enables training of tensors under conditions where individual training would result in low-quality or null tensors. In addition these methods often outperform the individual training even with all of the available training data.


\section{Acknowledgements}
 
Thanks Daniel Fried for the initial code and discussion, Laura Rimell and Massimiliano Pontil for
helpful discussions and comments. Supported by ERC Starting Grant DisCoTex (306920).
\clearpage
\bibliographystyle{myplainnat}
{\fontsize{10}{12}\selectfont
\bibliography{refs}
}
\end{document}